\documentclass[runningheads]{llncs}

\usepackage{eccv}

\usepackage{eccvabbrv}
\usepackage{caption}

\usepackage{graphicx}
\usepackage{booktabs}

\usepackage[accsupp]{axessibility}  
\usepackage{subcaption}

\usepackage{hyperref}

\usepackage{orcidlink}

\def\1n{\mathbf{1}_n}
\def\0{\mathbf{0}}
\def\1{\mathbf{1}}

\def\P{{\bf P}}

\definecolor{pink}{rgb}{0.9,0.5,0.5}
\definecolor{purple}{rgb}{0.5, 0.4, 0.8}   
\definecolor{gray}{rgb}{0.3, 0.3, 0.3}
\definecolor{mygreen}{rgb}{0.2, 0.6, 0.2}

\definecolor{greena}{rgb}{0.4, 0.5, 0.1}

\definecolor{bluea}{rgb}{0, 0.4, 0.6}

\definecolor{reda}{rgb}{0.6, 0.2, 0.1}

\newcommand{\cm}[1]{}

\newcommand{\myheading}[1]{\vspace{0.5ex}\noindent \textbf{#1}}

\definecolor{revcolor}{rgb}{0.0,0.34,0.72}

\definecolor{shocolor}{rgb}{0.80,0.0,0.0}

\newcommand{\Sref}[1]{Sec.~\ref{#1}}

\begin{document}

\title{Back-Tracking from Clarity: \\ Self-Learning to See Text from Afar} 

\titlerunning{Self-Learning to See Text from Afar}

\author{Duc Tri Tran\inst{1}\orcidlink{0009-0001-2185-8933} \and
Phi Le Nguyen\inst{1}\orcidlink{0000-0001-6547-7641}* \and
Minh Hoai\inst{2}\orcidlink{0000-0002-2415-6048
}}

\authorrunning{D.T. Tran et al.}

\institute{Institute for AI Innovation \& Societal Impact, Hanoi Univ. of Science \& Technology \and
 Australian Institute for Machine Learning, Adelaide University}

\maketitle

{\renewcommand{\thefootnote}{}\footnotetext{\textsuperscript{*}\,Corresponding author.}}

\begin{abstract}
  We propose a self-supervised framework designed to enhance the capability of scene text detectors in identifying and recognizing text in scenarios where instances are shown at significant distances, typically small, blurred, and frequently missed by conventional models. Our approach leverages the high-fidelity performance of existing text spotting models on large, clear text as a foundational supervisor. By temporally back-tracking these high-confidence detections through video sequences, we automatically synthesize pseudo-labels for preceding frames where the distant text is still visually degraded or undersized.
These pseudo-labels enable training a student model specialized for early text detection, without requiring any manual annotation. The success of this approach depends on accurate pseudo-label generation, for which we develop a dedicated scene text tracker capable of maintaining consistent text identities across challenging video sequences. In addition, we propose SceneText50, a diverse multilingual outdoor dataset to facilitate training and evaluation. Experiments show that our framework significantly improves early detection accuracy and robustness across varied scenes and languages. Code and data are at \href{https://github.com/trid2912/BackTrackingText}{https://github.com/trid2912/BackTrackingText}.

\end{abstract}

\section{Introduction}
\label{sec:intro}
The ability to detect and read scene texts from a distance is vital for safety-critical and time-sensitive applications \cite{zhang2021character, chen2004automatic}. In autonomous driving \cite{yuan2025piftext}, for instance, the early recognition of distant traffic signs, lane markings, or warning messages allows vehicles to anticipate hazards well in advance. Similarly, in robotics and intelligent navigation \cite{raisi2022text, hassan2025attention}, reading directional signs, door labels, or equipment identifiers from afar is essential for efficient path planning and task execution. For assistive technologies, such as wearable devices for the visually impaired, early detection of text at a distance provides users with critical, real-time contextual cues. Despite these needs, existing scene text detectors often struggle when text appears small, blurred, or distant. Their inability to process such challenging instances not only delays perception but also undermines the reliability of downstream reasoning, predictive planning, and overall safety.

The most straightforward strategy to address this is to train a detector using a comprehensive dataset of text across all scales. However, this approach is fundamentally constrained by the high cost and labor-intensive nature of large-scale manual annotation. Furthermore, simply increasing the amount of training data is often insufficient; without explicit focus, models frequently fail to generalize to the specific nuances of small, degraded text, while overly aggressive data augmentation can introduce noise that degrades overall performance. Moreover, the optimal threshold for distinguishable text varies significantly across environments, camera configurations, and linguistic contexts, rendering static, pre-trained models inherently inflexible when deployed in novel domains with distinct visual clutter or signage styles.

To bridge this gap, we propose a novel self-supervised learning framework that enhances small-text detection without the need for manual labels. Our paradigm leverages video sequences where text naturally transitions from small, blurry, and distant instances to larger, clearer ones. By employing a standard detector to identify confident, large-sized text instances and then using a dedicated tracker to propagate these detections backward through time, we automatically generate high-quality pseudo-labels for distant text. These labels allow the detector to iteratively extend its reach and sensitivity to smaller text instances.

Implementing this vision presents two primary hurdles: (1) the lack of robust trackers for distant text and (2) the absence of video datasets capturing these specific distance-varying dynamics. Existing generic object trackers \cite{Zhang2021ByteTrackMT, Chen2023SeqTrackST} are typically designed for objects with distinct, consistent appearance patterns. They struggle with scene text, which often features repetitive textures and high sensitivity to perspective and illumination. Moreover, these trackers generally operate via a track-by-detection paradigm, which is ill-suited for our backward-propagation requirements where distant instances lack reliable initial detections. Consequently, we require a propagation-based approach that maintains geometric consistency across frames, a task complicated by the fact that small, distant text is often textureless, blurred, or occluded.

To address these challenges, we make three major contributions:
\begin{itemize}
    \item \textbf{A specialized scene text tracker:} We develop a novel tracking mechanism optimized for backward propagation. Rather than tracking the text directly, which is unreliable at distance, we track the parent object ``carrier'' (e.g., a signpost or vehicle body) which offers more stable textural cues. We integrate this with a generic off-the-shelf tracker such as SAM-2 \cite{ravi2024sam}, and introduce a core polygon-based correspondence algorithm. This algorithm approximates each segmented region using a core polygon, allowing us to maintain geometric consistency and infer precise text box coordinates across frames via homography, even when the text itself is degraded.
    \item \textbf{A distance-varying video dataset:} We introduce a new dataset curated for distant scene text detection. Comprising 50 videos and over 440,000 annotated instances across diverse outdoor environments and languages, this dataset provides natural transitions from distant to close-range views. It serves as a necessary benchmark for evaluating the temporal consistency and early-detection capabilities of future models.
    \item \textbf{A self-supervised learning paradigm:} We introduce a novel framework that utilizes back-tracked pseudo-labels to train more robust detectors. By leveraging naturally occurring scale and clarity variations in videos, our proposed method learns to recognize distant text while maintaining peak performance on clear, large-scale instances, providing a scalable path towards more resilient scene text spotting systems.
\end{itemize}

\section{Related work}
\label{sec:related_work}

\myheading{Scene text detector.} Scene text detection has progressively evolved from detecting horizontal text using rectangular bounding boxes~\cite{lucas2005icdar, shahab2011icdar, karatzas2013icdar} to handling multi-oriented and irregular text through rotated boxes and quadrilateral representations~\cite{zhang2016multi, nagy2011neocr, zhou2015icdar, shi2017icdar2017}. Despite these advances, most detectors remain biased toward medium and large text instances, as reflected in common benchmarks such as ICDAR and Total-Text~\cite{karatzas2013icdar,zhou2015icdar,ch2020total}. Detecting distant or small text remains challenging due to limited spatial resolution and degraded appearance, a problem related to small-object detection~\cite{Nikouei_2025}. Existing solutions, including super-resolution approaches~\cite{li2017perceptual} or heavy data augmentation, attempt to improve recognition of small objects but often fail to capture the natural scale progression of objects across video frames.

\myheading{Scene text tracking.} Scene text tracking has received comparatively less attention. Recent works employ transformer-based video text spotting frameworks~\cite{wu2024end} or trackers built on strong image-based spotting models~\cite{he2024gomatching,ye2023deepsolo}. However, these approaches rely heavily on accurate detections and struggle when text appears very small or visually degraded. Detection-free tracking methods, such as Siamese-based trackers~\cite{bertinetto2016fully, li2019siamrpn++}, transformer-based trackers~\cite{cui2022mixformer}, or point-based trackers~\cite{karaev2025cotracker3}, alleviate this dependency but remain sensitive to appearance corruption and template drift. To overcome these limitations, we leverage the capabilities of SAM-2~\cite{ravi2024sam} to track the carrier object of the text and subsequently recover geometric correspondence across frames.

\myheading{Scene text datasets.}While several scene text datasets exist~\cite{ch2020total, karatzas2013icdar, zhou2015icdar, wu2021bilingual, nguyen2021dictionary, gupta2016synthetic, reddy2020roadtext, wu2024dstext}, they predominantly focus on clearly visible text and provide limited annotations for extremely small or distant instances. Additionally, these datasets fail to capture the core challenge of early text detection, where text must be annotated from very small and visually degraded states.

\myheading{Self-improved visual perception.} Recent work has explored \emph{self-improvement} strategies that enable perception systems to improve themselves without requiring additional human annotation by exploiting reliable predictions as supervision~\cite{lee2013pseudo,xie2020noisy}. For example, scene-adaptive object detection uses high-confidence detections together with tracked object trajectories to generate pseudo-labels for adapting an object detector to a new environment~\cite{m_Zhang-Hoai-CVPR23,m_Zhang-etal-BMVC24}. Similarly, in early event detection, annotations available for complete events have been used to train models that recognize the same event at an earlier stage~\cite{m_Hoai-DelaTorre-CVPR12,m_Hoai-DelaTorre-IJCV14,m_Tran-etal-ICIP21a}, while in action understanding, action recognition models have been used to supervise action anticipation networks without requiring separate anticipation annotations~\cite{m_Zhang-etal-BMVC21,m_Tran-etal-ICIP21b}. Our work follows the same self-improvement principle but addresses a fundamentally different problem. Rather than adapting to a new scene or anticipating future events, we exploit reliable detections of large, clear text and back-track them to generate supervision for small, distant text, thereby extending the detector’s capability to read text from afar without additional human annotation.

\section{Self-Learning Framework}
\begin{figure}[t]
    \centering
    \includegraphics[width=\linewidth]{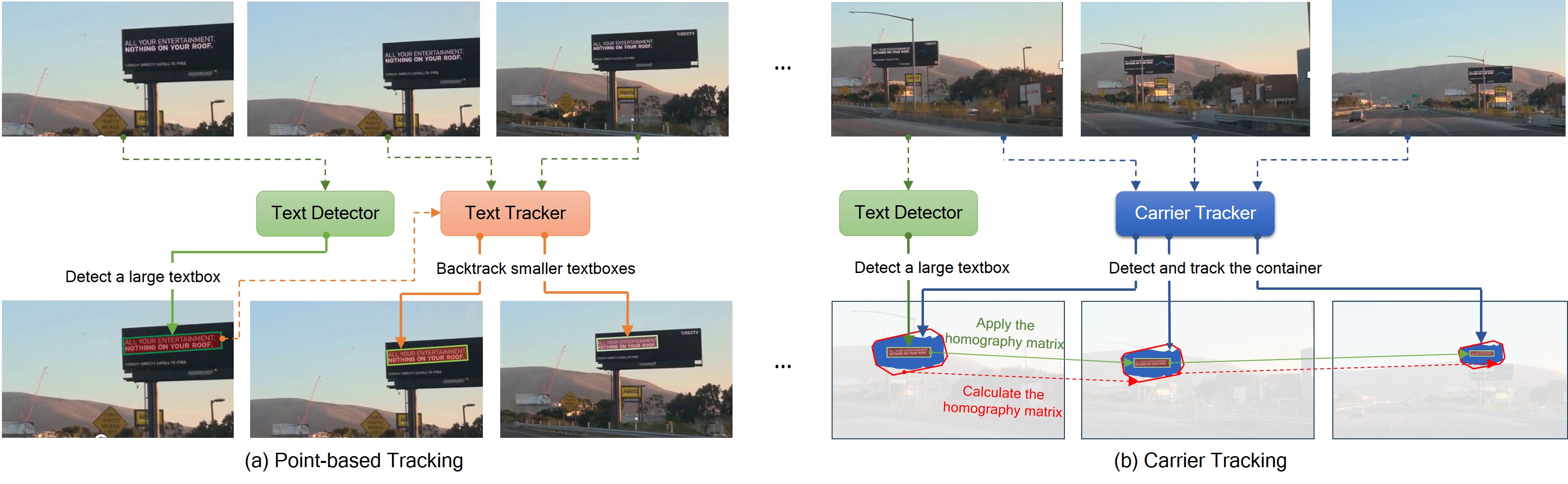}
    \caption{\textbf{Our algorithm generates pseudo-labels for small text instances.} A large text box is detected and tracked to infer smaller text instances. When texts are too small to track directly, we detect and track their carriers instead. A core polygon is extracted for each carrier, and a homography between consecutive frames is estimated to backtrack small-sized texts.}
    \vspace{-2em}
    \label{fig:main}
\end{figure}
This section presents our self-learning framework for training a scene text detector capable of reading text from afar. The key idea is leveraging a reliable existing detector to automatically generate supervision for smaller, more challenging ones. With sufficiently confident detection of large text in later frames, the model back-tracks these instances through the video to identify their earlier, smaller, and blurrier counterparts. These back-tracked samples serve as pseudo-labels for training a new distant-reading text detector without requiring any manual annotation, as shown in \cref{fig:main}.

\subsection{Large-to-small Tracking of Text Instances}

The ability to detect large text instances and back-track them to earlier frames where they appear smaller and blurrier is the key component of our framework. For detecting large scene text instances, any reliable text detector can be used. 
In this work, we employ DeepSolo~\cite{ye2023deepsolo} to produce text masks in our experiments, as its published model has been pretrained on multiple datasets. However, our framework is independent of detector choice and can be applied with any model capable of reliably detecting large, clear texts.

Once confident detections are obtained, the goal is to trace each text instance backward through earlier frames to recover its smaller and blurrier version. This process, referred to as \textit{large-to-small tracking}, forms the foundation for generating high-quality pseudo-labels used in the subsequent unsupervised training stage. The tracking procedure must satisfy two requirements: (1) it must robustly follow text instances across significant changes in scale, perspective, and visibility, and (2) it must maintain approximate geometric correspondence across frames to preserve each instance's orientation and character order. 

To achieve this, we develop a dedicated scene text tracker that leverages a general-purpose tracker to follow the object on which the text is printed and then employs a core polygon-based algorithm to establish correspondences between freeform segments across frames. The boundary of the text instance in the target frame is then recovered by relating the text region and its core polygon in the reference frame to the corresponding polygon in the tracked frame. 

\myheading{Problem Formulation.}  
Let $I_t$ denote an image frame at time step $t$, where a scene text instance is detected (e.g., by DeepSolo) and represented by a quadrilateral 
$\P_t = (P_t^1, P_t^2, P_t^3, P_t^4)$, with vertices ordered clockwise. 
The objective is to back-track this instance to the previous frame $I_{t-1}$ by estimating its corresponding quadrilateral 
$\P_{t-1} = (P_{t-1}^1, P_{t-1}^2, P_{t-1}^3, P_{t-1}^4)$. 

\myheading{Tracking the carrier object.} Unfortunately, directly tracking the text instance itself is highly unreliable, especially when the text becomes small, blurred, or partially occluded in earlier frames. Scene text often exhibits weak texture and repetitive patterns, making feature matching unstable. 
Our idea to overcome these limitations is to track the carrier object on which the text is printed, such as a signboard or vehicle panel, instead of the text pixels directly. These carrier objects generally provide richer visual cues and more stable boundaries, allowing for robust tracking with generic object trackers, after which the text region can be localized through geometric correspondence within the tracked segment.

Another advantage of the proposed approach is that it allows us to leverage a powerful general-purpose object tracker for our task. In this work, we adopt SAM-2~\cite{ravi2024sam}, a state-of-the-art model for prompt-based tracking.
We use a single-point prompt placed at the center of the detected text instance $\P_t$ to initiate tracking. Given this prompt, SAM-2 produces segmentation masks $M_t$ and $M_{t-1}$ corresponding to the same prompted point in frame $I_t$. These masks represent the segmented regions of the carrier object associated with the text instance. However, the segmentation masks are not necessarily quadrilateral or perfectly aligned with the text region we aim to track. Moreover, while $M_t$ and $M_{t-1}$ correspond to the same object, SAM-2 does not provide dense pixel-level correspondences or boundary alignment between the two masks, which are essential for accurately inferring the quadrilateral $\P_{t-1}$ of the tracked text instance.

\begin{figure}[t]
    \centering
    \includegraphics[width=\linewidth]{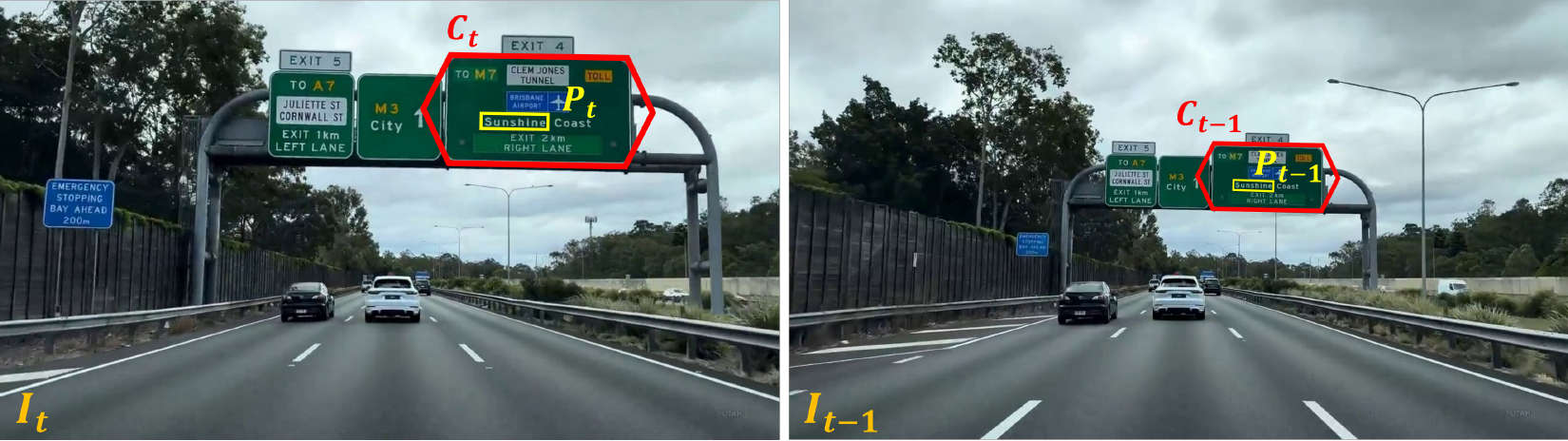}
    \caption{Geometric correspondence through core polygon at frame t (left) and t-1 (right) in case n=6. The mapping $\mathbf{H}$ between core polygons $C_t$ and $C_{t-1}$ is used to project the detected text region $P_t$ back to its position in the previous frame $P_{t-1}$.
    \vspace{-1em}
    \label{fig:geometric_correspond}}
\end{figure}

\myheading{Geometric correspondence through core polygons.}
To recover geometric correspondence between text instances across frames, we first fit two $n$-vertex core polygons, $C_t$ and $C_{t-1}$, to the segmentation masks $M_t$ and $M_{t-1}$, respectively (the definition of a core polygon and its fitting procedure will be described in the next paragraph). Each core polygon serves as a convex approximation of the free-form mask boundary, providing a structured and parametric representation of the region. We then use the Random Sample Consensus (RANSAC) algorithm~\cite{fischler1981random} to estimate a homography matrix $\mathbf{H}$ that maps $C_t$ to $C_{t-1}$. This homography captures the geometric transformation between the two carrier objects and is applied to project the detected text region $\P_t$ in frame $I_t$ to its estimated position $\P_{t-1}$ in frame $I_{t-1}$, as shown in \cref{fig:geometric_correspond}. The resulting $\P_{t-1}$ preserves the geometric structure and orientation of the text instance while accounting for perspective and motion changes.

\myheading{Core polygon estimation.} 
One approach to parameterizing a free-form segmentation mask is to utilize its convex hull. However, the number of hull vertices inherently varies across different instances and frames, making them unsuitable for establishing consistent geometric correspondence in video sequences.To address this limitation, we propose an algorithm to approximate the segmentation mask using a polygon with specific structural properties. Intuitively, our approximation satisfies three criteria: (1) the number of vertices is fixed; (2) the orientations of the polygon edges are predefined and constant; and (3) the approximation error between the polygon and the mask is as small as possible. The first two properties are designed to establish a consistent topology between core polygons across successive frames, while the third ensures that the spatial relationship between the core polygon and the enclosed text remains stable over time. Formally, for any $n$-gon, we define an $n$-core polygon $\mathcal{P}_n$ as a polygon characterized by the following properties:

\begin{definition}
    For an integer $n \geq 3$, a core polygon $\mathcal{P}$ of a mask is a polygon satisfying:
    \begin{itemize}
        \item It is a circumscribing polygon of the mask (meaning its edges are tangent lines to the mask), and its number of vertices does not exceed $n$.
        \item All of its internal angles (excluding those coincident with the mask's convex hull) are equal to $\frac{(n-2)\pi}{n}$.
    \end{itemize}
    \label{def:core_polygon}
\end{definition}

\begin{figure}
    \centering
    \includegraphics[width=\linewidth]{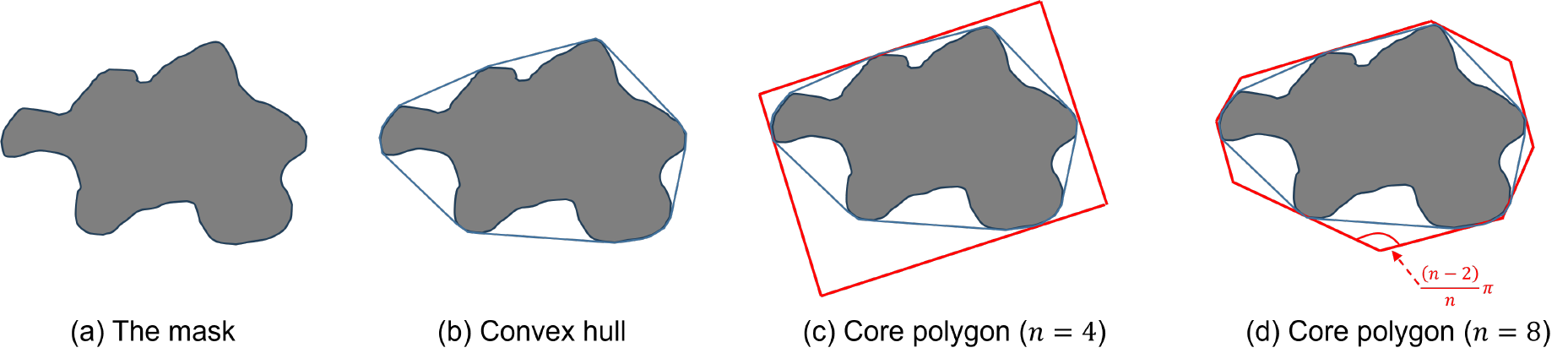}
\vskip -0.1in
    \caption{\textbf{Illustration of core polygons.} A core polygon is the smallest equiangular polygon that circumscribes the mask.
By increasing the number of vertices of the core polygon, the distance between the core polygon and the convex hull of the mask can be made arbitrarily small.}
\vspace{-3em}    
    \label{fig:core_polygon}
\end{figure}

In the following, we demonstrate that by increasing the number of vertices $n$, the approximation error between the core polygon and the mask vanishes, with their ratio converges to $1$. \Cref{fig:core_polygon} illustrates the effectiveness when increase the number of vertices $n$.
\begin{theorem}
Let $\mathcal{Q}$ be a convex polygon, and let $\mathcal{P}$ be its circumscribing polygon. Denote $P_1, P_2, \dots, P_n$ as the vertices of $\mathcal{P}$ (ordered in the counterclockwise direction). Assume that $P_{r_1}, \dots, P_{r_m}$ are the vertices of $\mathcal{P}$ that are not vertices of $\mathcal{Q}$. 
We define the Looseness Measure ($\mathcal{L}$) as the ratio of the perimeter of the circumscribing polygon to the perimeter of the convex polygon. Therefore, the Looseness Measure is bounded by

\begin{equation}
\mathcal{L} = \frac{p_{\mathcal{P}}}{p_{\mathcal{Q}}} \leq \frac{1}{\sin \frac{\rho}{2}},
\end{equation}
where $\rho = \min \widehat{P}_{r_i}$ is the minimum internal angle of the core polygon.
\end{theorem}
For a core polygon, all interior angles are equal, so $\rho = \frac{(n-2)\pi}{n}$. As the number of vertices $n$ increases, this constant interior angle approaches $\pi$. Consequently, the ratio between the perimeter of the core polygon and that of the mask’s convex hull converges to $1$. 

\subsection{Auto-generated labels and model training}

This stage uses the large-to-small tracking framework to generate training labels for distant text and to train a detector capable of reading text from afar.

\myheading{Label generation.} 
To generate high-quality training labels, we employ a multi-stage pipeline that prioritizes high-confidence detections while ensuring temporal continuity through advanced tracking. The process is initialized using a reliable large-text detector to identify primary text instances. To expand these labels and maintain temporal consistency, we perform back-tracking using two ways:
\begin{enumerate}
    \item Point-based Tracking: We first utilize CoTracker3~\cite{karaev2025cotracker3} to track the scene text by its bounding points across preceding frames. To ensure label integrity, we only retain tracking results that maintain a confidence score $> 0.8$.
    \item Carrier Object Tracking: For instances reaching the final frame of the sequence, we further backtrack by identifying the underlying carrier object (the surface or item the text is attached to) using SAM-2~\cite{ravi2024sam}.
\end{enumerate}
To resolve potential spatial conflicts, since detections occur independently per frame, we implement a strict priority protocol. Detection masks are prioritized; a back-tracked mask is integrated into the final label set only if it does not significantly overlap with an existing detection. This hybrid approach, combining point-level precision with object-aware segmentation, produces a diverse and dense set of pseudo-labels spanning a wide range of text scales, distances, and motion profiles. Note that this multi-module tracking pipeline and the use of future frames occur \emph{only} during offline pseudo-label generation. Once trained, the student detector runs independently on each frame, without future frames or any tracking module, incurring no additional inference overhead.

\myheading{Training.}
We freeze the image encoder, and train the decoder modules using the auto-generated pseudo-labels. To ensure stable convergence and better generalization, we employ Exponential Moving Average (EMA) during training. This adjustment allows the model to capture the fine-grained geometry of small and distant text more effectively than standard sampling densities.
All experiments use a single NVIDIA RTX A6000 (48GB), AdamW optimizer, weight decay $10^{-4}$, EMA decay 0.999 and batch size 8.

\section{SceneText50 dataset}

\begin{figure}[t]
    \centering
    \includegraphics[width=\linewidth]{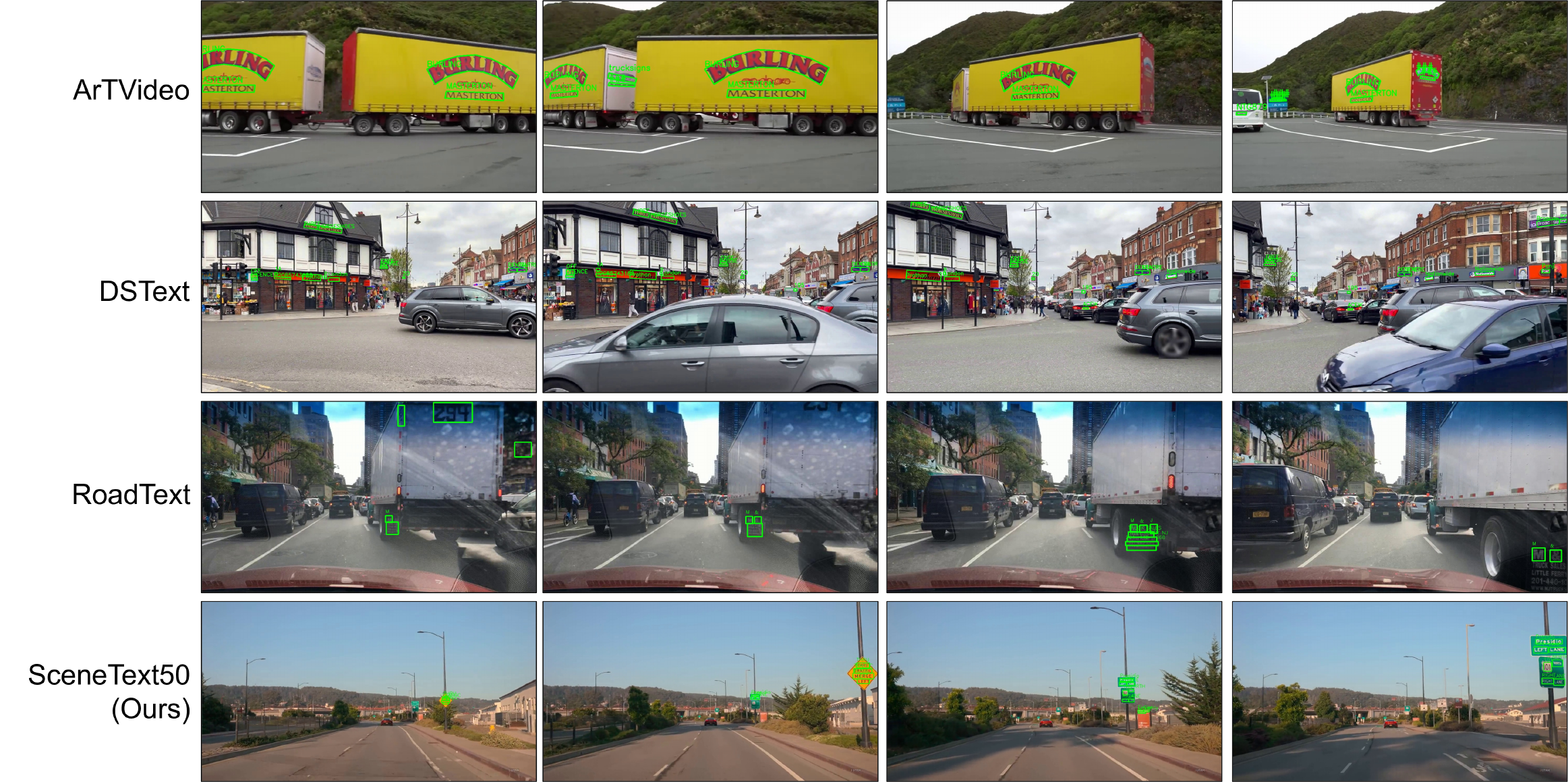}
    \vskip -0.05in
    \caption{\textbf{Examples of text instances in ArTVideo, DSText, RoadText, and SceneText50 datasets.} 
    While ArTVideo focuses on large text and often ignores smaller instances, DSText contains small text that remains largely unchanged in size throughout the sequence. RoadText frequently merges labels of different text together, whereas our SceneText50 dataset uniquely contains small-to-large text tracks.}
    \label{fig:data_dif}
\end{figure}

Developing and evaluating methods for distant text detection requires video data in which scene text appears across a wide range of scales-from far, small, and blurry to near and fully legible-within the same trajectory. However, existing video datasets for scene text (\eg, ArTVideo~\cite{he2024gomatching}, DSText~\cite{wu2024dstext}, RoadText~\cite{reddy2020roadtext}) largely feature text of relatively consistent size within each sequence, offering limited scale variation and providing few examples of genuinely small or distant text, as shown in \cref{fig:data_dif}. Such datasets are insufficient for studying large-to-small back-tracking, for training models capable of reading text from afar, and for reliably evaluating performance on small text. To address this gap, we collect a new dataset, SceneText50, designed specifically to contain size-varying text instances within their tracks, including many instances of small and distant text. This section describes the data collection process, annotation protocol, and key statistics of SceneText50.

\myheading{Video data collection.}
To support distant text detection, our dataset must contain video sequences in which scene text appears across substantial scale changes, from small and barely legible at a distance to large and clear as the camera approaches. Such scale variation naturally occurs in scenarios where a camera is worn or mounted on a moving platform (\eg a person, a robot, or a car). Motivated by these application settings, we specifically search for ego-centric walking and driving videos in which the camera moves through real environments and text gradually becomes larger as the viewer approaches it.

To find such videos, we use YouTube search queries including ``4K walking'', ``video walking'', ``video driving'', and ``4K driving'', which return many ego-centric footages with sufficient resolution and forward motion for our task. Although our self-learning framework is not limited to any particular language or script, in this work we focus on scenes containing Latin-alphabet text. This choice is driven by practical considerations: (1) collecting and annotating data across all language families would exceed our available resources, and (2) Latin-script text spans many different languages and is broadly used worldwide, making it a representative and impactful starting point. Importantly, our approach generalizes to non-Latin scripts as well, and future extensions of SceneText50 may include additional languages and regions. Following this data collection procedure, we gather 50 videos, each 1-2 hours long, from 17 different countries across multiple continents (North America, South America, Europe, Asia, Oceania, and Africa).

\myheading{Data division and annotation.} 
After collecting the videos, we manually divide each long video into short clips of 5-15 seconds. Each clip is selected to contain at least one scene text instance that appears small at first and gradually becomes larger as the camera approaches. For a typical video, this process yields approximately 100-200 clips. From each video, we then uniformly sample about 10\% of these clips and aggregate them into our final dataset, resulting in a total of 841 video clips of 5-15 seconds each. We instruct annotators to label every scene text instance appearing in these 841 clips using quadrilateral bounding boxes. Importantly, these annotations are used only for evaluation; our self-learning framework does not require human annotation for training.

To construct the train/test split, we divide the labeled clips of each video based on their temporal order: the first two thirds are designated as training data, and the remaining one third as test data. This split is designed to simulate the real application scenario of on-the-fly self-learning in a new environment. For example, consider a robot navigating a new building or a person wearing smart glasses walking through a new city. As they move through the environment, the system has time to observe and adapt through self-learning. After this adaptation period, the system should be able to detect small, distant text more effectively in the same general environment, such as the same city, building, or visual style, which are not necessarily the exact same corridor for a robot or the identical home-to-work route for a person. Our dataset design reflects this intended scenario: the first two-thirds and the last one-third of each video naturally share the same broad environment while still offering different scenes, viewpoints, and lighting conditions.

We use only the ground-truth annotations from the test clips to evaluate the performance of the distant text detector, ensuring that model training remains annotation-free while enabling rigorous and unbiased performance assessment.

\myheading{Data statistics.}
SceneText50 contains a total of 67,371 annotated frames, forming 14,263 unique text trajectories. Each trajectory captures a text instance as it transitions from one size and perspective to another. Across all trajectories, the minimum and maximum text areas average 288 and 1,288 pixels respectively, yielding an average scale ratio of 7.94. This large variation, far greater than what is typically seen in existing datasets, indicates that many trajectories undergo substantial scale changes, which is essential for evaluating distant text detection.

\begin{figure}[t]
\centering

\begin{minipage}[t]{0.56\textwidth}
\centering
\captionof{table}{\textbf{Statistics of SceneText50 and comparison with other text video datasets.}}
\label{tab:statistic}
\resizebox{\linewidth}{!}{%
\begin{tabular}{lrrlrr}
\toprule
& & & & \multicolumn{2}{c}{\textbf{Text size}} \\
\cmidrule{5-6}
\textbf{Dataset} & \textbf{\#Video} & \textbf{\#Frame} & \textbf{Annotation} & \textbf{Average} & \textbf{Variance} \\
\midrule
ArTVideo \cite{he2024gomatching} & 60 & 14.2K & Polygon & 3,539 & 59016 \\
DSText \cite{wu2024dstext} & 140 & 62.1K & Rectangle & 1,758 & 15769 \\
RoadText \cite{reddy2020roadtext} & 1000 & 300K & Upright rectangle & 2,141 & 24406 \\
SceneText50 & 841 & 67.3K & Quadrilateral & 985 & 340411 \\
\bottomrule
\end{tabular}%
}
\end{minipage}
\hfill
\begin{minipage}[t]{0.40\textwidth}
\centering
\vspace{0pt}
\includegraphics[width=\linewidth]{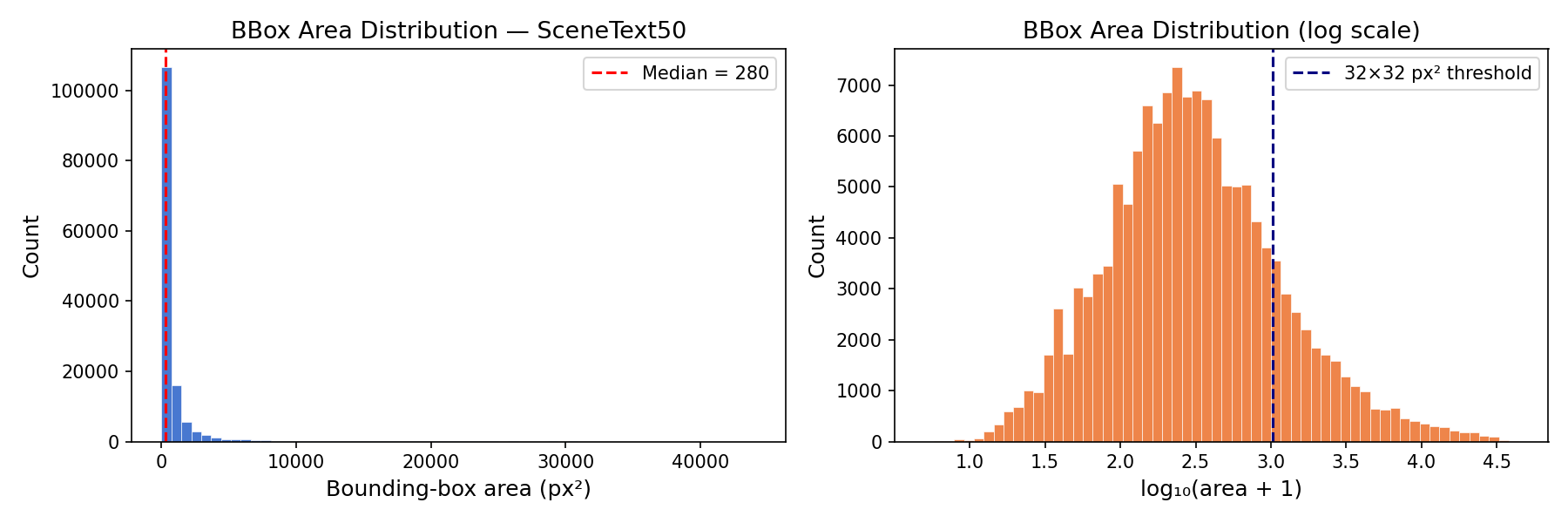}
\captionof{figure}{Distribution of bounding-box areas in SceneText50, skewed toward small text instances.}
\label{fig:bbox_distribution}
\end{minipage}

\vspace{-1.5em}
\end{figure}

\Cref{tab:statistic} compares SceneText50 with major video-based scene text datasets. ArTVideo~\cite{he2024gomatching} focuses on arbitrarily shaped text but contains relatively few videos. DSText~\cite{wu2024dstext} includes dense and small text but covers only six predefined scenarios. RoadText~\cite{reddy2020roadtext} provides large-scale street-view text but generally features text of relatively stable size across trajectories.

In contrast, SceneText50 offers two unique advantages. First, it explicitly targets videos where text appears at small sizes and grows larger as the camera approaches, which is a natural property of outdoor and ego-centric video that existing datasets largely overlook, as evidenced in \cref{fig:data_dif}. Second, it features the smallest average text area among all compared datasets, making it particularly well-suited for studying distant and small-scale text detection. Additionally, SceneText50 also exhibits a high text density per frame, surpassing even RoadText despite covering a broader range of environments. The distribution of bounding-box areas in \cref{fig:bbox_distribution} further confirms that the dataset is dominated by small text instances.

\section{Experiments}
\subsection{Evaluation settings}

\myheading{Tracking.} The tracking evaluation follows a back-tracking paradigm to maintain consistency across the training data. The process is initialized at the last frame of each sequence, where the ground-truth bounding box of the scene text is used as the initial template. The tracker then proceeds to estimate the object's trajectory in reverse, from the final frame back to the first frame. To ensure a fair comparison between various tracking architectures, all models are initialized using the same standardized bounding box coordinates. This setup minimizes the impact of initialization bias and focuses the evaluation on the tracker's ability to maintain spatial consistency and handle temporal degradation over the length of the video sequence.

\myheading{Self-learning to read from afar.} We investigate the efficacy of two distinct strategies for small text detection: Synthetic Data Augmentation and Pseudo-Label Generation. To rigorously evaluate these approaches, we establish two experimental tracks: \textbf{(1) Detector Consistency:} We train multiple detection architectures under both settings to evaluate their impact on standard detection metrics. Performance is measured by the average Precision and Recall across the testing set to determine which labeling strategy produces more discriminative feature representations. \textbf{(2) Early Spotting Performance:} We evaluate the model's capacity for early detection and early spotting (recognition) within a text-spotting framework. In this setting, the teacher and student models share an identical architecture to isolate the influence of the data-generation strategy. To quantify temporal efficiency, we measure the Average Detection Delay, defined as the number of frames elapsed between the initial appearance of a scene text instance, typically in a blurred, small state, and its first correct identification and recognition at a predefined IoU threshold.

\subsection{Experiment metrics \label{sec:metrics}}
To evaluate tracking performance, we follow LaSOT~\cite{fan2019lasot} and employ three primary metrics: Precision@20px, Success@0.5IoU, and Area Under Curve (AUC). Precision@20px measures the percentage of frames where the Euclidean distance between the tracked target's center and the ground truth is within a 20-pixel threshold. Success@0.5IoU represents the ratio of frames where the Intersection over Union (IoU) between the predicted and ground-truth bounding boxes exceeds 0.5. Finally, the AUC provides a comprehensive assessment of the tracker's overall robustness by calculating the area under the success curve across all IoU thresholds from 0 to 1.

To evaluate the performance of early detection, we utilize an evaluation framework that measures the trade-off between Average Detection Delay and the False Alarm Rate (FAR), following the protocol from \cite{lao2019minimum}. This approach defines a correct detection based on spatial accuracy-requiring an Intersection over Union (IoU) overlap above a specific threshold. FAR represents the proportion of total declarations that are incorrect, serving as a metric for the system's reliability. Meanwhile, Detection Delay quantifies temporal efficiency by calculating the number of frames elapsed between an object's first appearance and its initial detection, with a maximum delay penalty applied to any ground truth objects that the system fails to identify.

\subsection{Results}

\begin{table}[t]
\centering
\caption{Text tracking results on three datasets. The proposed method outperforms several state-of-the-art trackers across three performance metrics. Prec and Succ denote Precision@20px and Success@0.5IoU, respectively, as defined in \Sref{sec:metrics}.}
\label{tab:overall_result}
\small
\setlength{\tabcolsep}{2.5pt} 
\resizebox{\textwidth}{!}{\begin{tabular}{@{}l|ccc|ccc|ccc@{}}
\toprule
 & \multicolumn{3}{c|}{\textbf{SceneText50}} & \multicolumn{3}{c|}{\textbf{ArTVideo}} & \multicolumn{3}{c}{\textbf{DSText}} \\
\textbf{Models} & AUC $\uparrow$ & S@50 $\uparrow$ & P@20 $\uparrow$ & AUC $\uparrow$ & S@50 $\uparrow$ & P@20 $\uparrow$ & AUC $\uparrow$ & S@50 $\uparrow$ & P@20 $\uparrow$ \\
\midrule
SiamRPN++~\cite{li2019siamrpn++} & 30.24 & 29.62 & 57.53 & 80.65 & 76.64 & 63.58 & 61.82 & 48.15 & 42.16 \\
STARK~\cite{yan2021learning} & 37.55 & 39.99 & 63.80 & 83.65 & 73.25 & 67.53 & 64.17 & 52.12 & 46.40 \\
MixFormer~\cite{cui2022mixformer} & 37.77 & 38.39 & 66.48 & 86.58 & 80.52 & 72.02 & 63.58 & 52.39 & 46.76 \\
SeqTrack~\cite{Chen2023SeqTrackST} & 37.55 & 39.14 & 62.97 & 88.53 & 80.11 & 72.85 & 65.24 & 54.02 & 47.72 \\
ARTrackv2~\cite{bai2024artrackv2} & 36.08 & 37.28 & 61.59 & 85.84 & 73.50 & 67.20 & 66.06 & 52.78 & 48.61 \\
SAM-2~\cite{ravi2024sam} & 26.43 & 22.38 & 56.09 & 39.48 & 41.09 & 39.48 & 46.27 & 50.37 & 44.65 \\
CoTracker3~\cite{karaev2025cotracker3} & 38.65 & 44.01 & 57.02 & 90.99 & 85.75 & 76.74 & 80.59 & 63.59 & 57.54 \\
\textbf{Proposed} & \textbf{44.71} & \textbf{49.23} & \textbf{74.62} & \textbf{91.19} & \textbf{86.15} & \textbf{76.83} & \textbf{85.95} & \textbf{68.14} & \textbf{73.62} \\
\bottomrule
\end{tabular}}
\end{table}

\myheading{Tracking results}. \Cref{tab:overall_result} provides a quantitative comparison of the proposed text tracking method against several state-of-the-art trackers across the SceneText50, ArTVideo, and DSText benchmarks. Notably, on the SceneText50 dataset, the proposed method reaches a Precision of 74.62, representing a significant improvement over the 66.48 achieved by MixFormer. Furthermore, the method demonstrates robust generalization on the challenging DSText dataset, where it secures an AUC of 85.95 and a Precision of 73.62, substantially outperforming the next-best model, CoTracker3, which records 80.59 and 57.54 respectively.

\begin{figure}[t]
    \centering
    \includegraphics[width=1\linewidth]{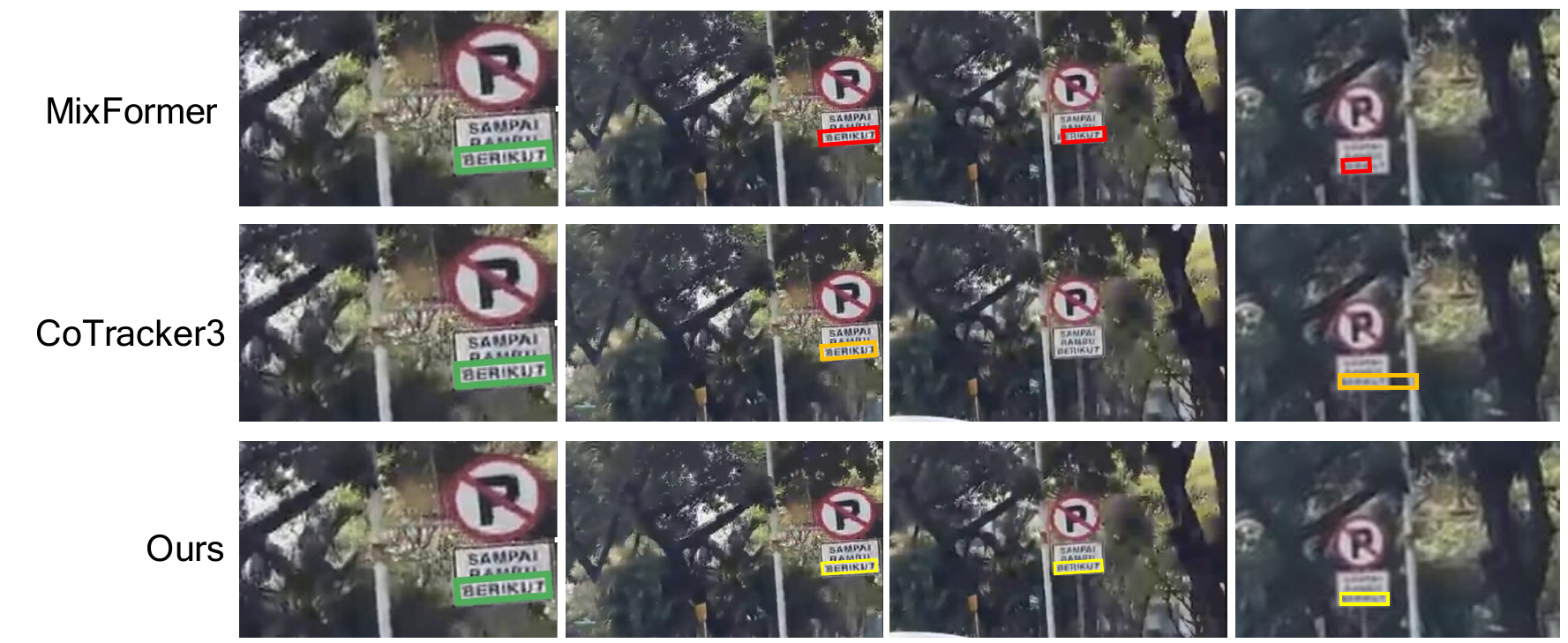}
    \vskip -0.1in
    \caption{\textbf{Qualitative comparison of back-tracking performance.} The sequence is initialized using the ground-truth bounding box at the final frame (left, green bbox) and propagated backward to earlier frames.
    Rows depict the trajectories generated by a single-object tracker (MixFormer), a point-based tracker (CoTracker3), and our proposed approach using carrier-object tracker, respectively.}
    \label{fig:qualitative_track}
\end{figure}

In addition, the qualitative results in \cref{fig:qualitative_track} demonstrate that traditional single-object trackers, such as MixFormer, often lose the target as it regresses toward the first frame, where the text is significantly smaller and blurred. Similarly, point-based trackers like CoTracker3 may suffer from point-drift when the local texture of the scene text becomes indistinguishable from the background. In contrast, the carrier-object approach leverages the larger, more stable context of the object carrying the text (the traffic sign in this case). This provides a more reliable spatial anchor, allowing the model to infer the text's position even when the characters themselves are no longer legible.

\begin{table}[t]
\centering
\caption{Comparison of different approaches for generating training data to improve the performance of a scene text detector, either using traditional data augmentation or our text-tracking results. We consider three types of detection models: an open-vocabulary object detector (Grounding DINO), a scene text detector (DPText-DETR), and a scene text spotter (DeepSolo). Training with tracked text leads to significantly higher recall while maintaining similar or higher precision.}
\label{tab:Detection}
\small
\setlength{\tabcolsep}{8pt} 
\resizebox{0.8\textwidth}{!}{\begin{tabular}{@{}lccc@{}}
\toprule

\textbf{Models} & \textbf{Training Data by} & \textbf{Precision} $\uparrow$ & \textbf{Recall} $\uparrow$  \\

\midrule
Grounding DINO~\cite{liu2024grounding} & Augmentation & 14.21 & 29.77  \\
 & Text-tracking (proposed) & {34.03} & {34.74}  \\
\midrule
DPText-DETR~\cite{ye2023dptext} & Augmentation & 44.61 & 21.80  \\
 & Text-tracking (proposed) & 44.58 & {23.12}  \\
\midrule
DeepSolo~\cite{ye2023deepsolo} & Augmentation & 52.25 & 37.75  \\
 & Text-tracking (proposed) & 51.88 & {40.27} \\

\bottomrule
\end{tabular}}
\vspace{-2.5em}
\end{table}

\myheading{Results on self-learning to detect and spot text from afar.}
\Cref{tab:Detection} presents a comparative analysis of scene text detection and spotting performance when models are fine-tuned using two data strategies: our proposed pseudo-labeling framework (via backtracking) and a traditional data augmentation approach. For the latter, we randomly sample five frames from each video in the ArTVideo~\cite{he2024gomatching} and DSText~\cite{wu2024dstext} datasets, resize them to half of their original size, and apply Gaussian blur to synthesize small-text instances. This process produces a synthetic augmentation dataset consisting of 1,000 images.

Across three distinct architectures, our method consistently yields higher Recall compared with the standard augmentation strategy. This effect is most pronounced for the open-vocabulary Grounding DINO model, where pseudo-labeling more than doubles the Precision from 14.21 to 34.03 and improves Recall from 29.77 to 34.74. These results suggest that the proposed pseudo-labeling method captures a more comprehensive set of text instances, particularly small or blurred ones, thereby improving the model’s sensitivity in scene text detection and spotting tasks.

\begin{figure}[t]
    \centering

    \begin{subfigure}[h]{0.45\linewidth}
        \centering
        \includegraphics[width=\linewidth]{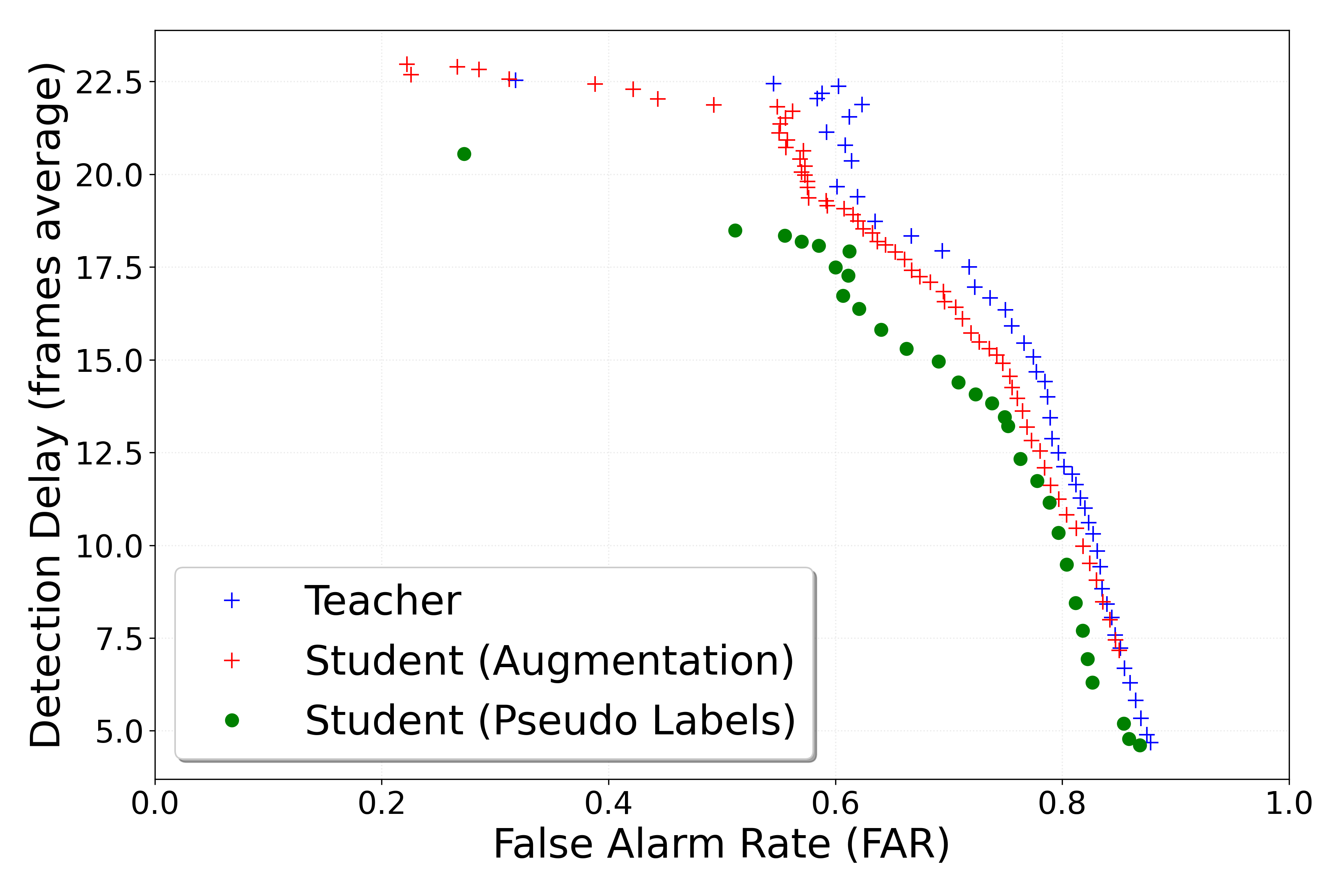
        }
        \caption{Text detection}
        \label{fig:det_delay}
    \end{subfigure}
    \hfill
    \begin{subfigure}[h]{0.45\linewidth}
        \centering
        \includegraphics[width=\linewidth]{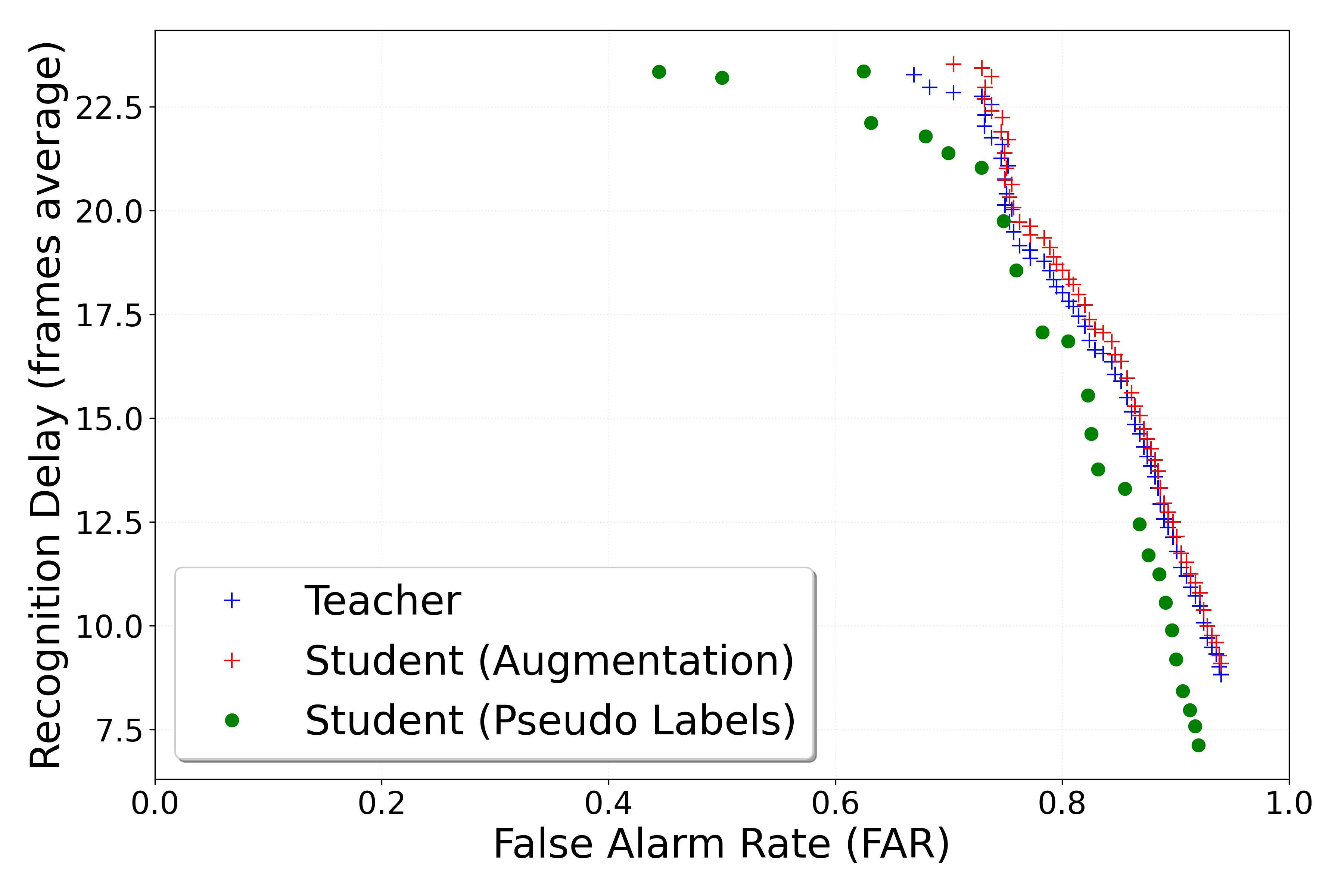}
        \caption{Text detection and recognition}
        \label{fig:reg_delay}
    \end{subfigure}
    \vskip -0.1in
\caption{\textbf{Evaluation of early detection and early spotting performance.} 
The plots illustrate the trade-off between the False Alarm Rate (FAR) and the average Detection Delay (measured in frames) for both (a) detection-only and (b) end-to-end spotting tasks. We compare the Teacher model (without self-learning) against Student models trained with additional data, either using traditional data augmentation or our proposed pseudo-labeling pipeline. Across both tasks, the student model trained with pseudo-labels (green) 
demonstrates a stronger ability to identify and recognize scene text at earlier stages of its appearance.}
    \label{fig:far_delay}
\end{figure}

Furthermore, \Cref{fig:far_delay} demonstrates the efficacy of the proposed pseudo-labeling strategy in enhancing the temporal efficiency of scene text systems. In both detection-only and end-to-end spotting scenarios, the pseudo-labeling approach shifts the performance frontier toward the lower-left quadrant, indicating that the model can correctly identify and recognize text instances significantly earlier in a video sequence for any given reliability level. This enhanced capability suggests that the backward-propagated pseudo-labels provide more discriminative features for small, blurred, and distant text, allowing the detector to trigger a correct declaration closer to the object’s first appearance frame.


\Cref{fig:early_infer} illustrates the superior sensitivity of the pseudo-labeling strategy in long-range scenarios. While the Teacher lacks the discriminative power to detect the distant sign, the Student (Augment) demonstrates improved localization but fails at accurate recognition. In contrast, the Student (Pseudo-labels) model achieves correct end-to-end spotting. This validates that training on backward-propagated labels provides more authentic representations of text degradation, directly supporting the reduced detection delays observed in our quantitative temporal analysis.
Note that detection delay measures \emph{observational latency}, i.e., how early the system can first detect and read text, rather than \emph{processing latency} (per-frame inference time). Since our method does not change the detector architecture, processing latency is identical to the base student model, while the observational latency it reduces is what matters for early reading from afar.

\myheading{Ablation studies.} 
In the setting of carrier-based tracking only, increasing the number of vertices generally improves all metrics: AUC rises from 32\% (4 vertices) to 34.6\% (16-20 vertices), Success@0.5IoU
increases from 33\% to 36\%, and Precision@20px improves from 56\% to 58\%. We observe a distinct performance plateau beyond the 16 vertex mark.

The effectiveness of using a carrier object is shown in \cref{tab:ablation_study_1}. 
Results show that combining point-based and carrier-based tracking components improves significantly across all metrics. This improvement indicates that carrier-based tracking effectively serves as a fallback mechanism when point-based tracking fails, helping recover difficult cases and leading to more robust and accurate overall tracking performance.

In addition, we evaluate the impact of the Core Polygon (CP) representation and the RANSAC-based refinement on the tracking performance, as shown in \cref{tab:ablation_study}.
Using Convex Hull with homography estimation (CH+H) achieves the lowest performance, and RANSAC refinement provides only marginal benefits. In contrast, when the Core Polygon representation is used together with RANSAC refinement (CP+R), the performance improves significantly 
, demonstrates that the use of CP plays a crucial role in homography matching for carrier tracking.

\begin{table}[htbp]
\centering
\begin{minipage}[!t]{0.495\columnwidth}
    \centering

    \includegraphics[width=\linewidth]{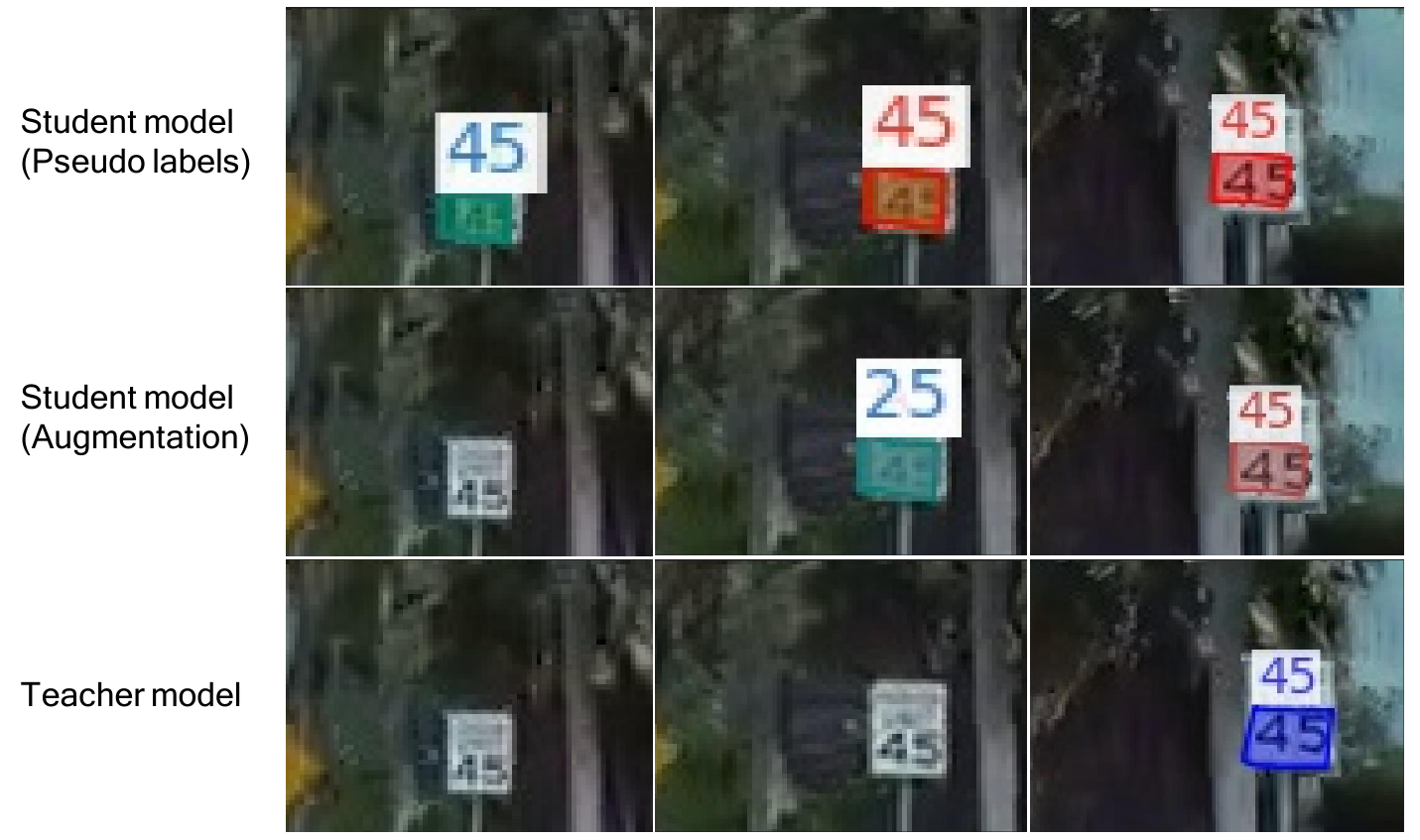}
    \captionof{figure}{Qualitative inference comparison of the teacher and student models on distant scene text.}
    \label{fig:early_infer}
    
\end{minipage}%
\hfill
\begin{minipage}[!t]{0.485\columnwidth}
    \centering
    \caption{Ablation study of tracking with and without carrier.}
    \vspace{-1em}
    \label{tab:ablation_study_1}
    \resizebox{0.85\linewidth}{!}{%
    \begin{tabular}{@{}lccc@{}}
    \toprule
    Tracking approach & AUC $\uparrow$ & S@0.5 $\uparrow$ & P@20 $\uparrow$ \\
    \midrule
    Point-based Tracking  & 38.62 & 44.23 & 57.44 \\
    Carrier-based Tracking  & 34.68 & 36.00 & 57.66 \\
    \textbf{Proposed (Full)} & \textbf{44.71} & \textbf{49.23} & \textbf{74.62} \\
    \bottomrule
    \end{tabular}}
    \vspace{0.5em}
    \caption{Impact of Core Polygon (CP) and RANSAC-based refinement, where CH, H, and R denote Convex Hull, Homography, and RANSAC, respectively.}
    \vspace{-1em}
    \label{tab:ablation_study}
    \resizebox{0.85\linewidth}{!}{%
    \begin{tabular}{@{}lccccc@{}}
    \toprule
    Method & Poly. & RAN. & AUC $\uparrow$ & S@0.5 $\uparrow$ & P@20 $\uparrow$ \\
    \midrule
    CH+H & $\times$      & $\times$      & 37.30 & 42.07 & 62.57 \\
    CH+R & $\times$      & $\checkmark$  & 37.40 & 42.10 & 62.61 \\
    CP+R & $\checkmark$  & $\checkmark$  & \textbf{44.71} & \textbf{49.23} & \textbf{74.62} \\
    \bottomrule
    \end{tabular}}
\end{minipage}
\end{table}

\section{Conclusion}

We have introduced a novel scene text tracker designed to address the problem of large-to-small scene text tracking through a hybrid approach that synergizes point tracking and carrier tracking. Evaluations across multiple scene text datasets demonstrate that our method consistently outperforms current state-of-the-art trackers. Furthermore, we leverage this framework to develop an automated labeling pipeline specifically for distant text instances. Experiments across multiple detectors show that fine-tuning on our auto-generated labels yields superior performance compared to standard augmentation strategies, significantly boosting recall and enabling the model to correctly identify and recognize text instances much earlier in video sequences. Finally, we introduce SceneText50, the first dedicated benchmark for distant-to-close scene text tracking, filling a critical gap in early text detection research.


\myheading{Acknowledgments.} This work is partially supported by VinUniversity under Grant No VUNI.2122.SG04. 

%
%
\bibliographystyle{splncs04}
\bibliography{longstrings,main}

@string{aaai = "Proceedings of AAAI Conference on Artificial Intelligence"}

@string{cvpr   = "Proceedings of the {IEEE} Conference on Computer Vision and Pattern Recognition"}

@string{eccv  = "Proceedings of the European Conference on Computer Vision"}

@string{icra =  "Proceedings of the International Conference on Robotics and Automation"}

@string{icml =  "Proceedings of the International Conference on Machine Learning"}

@string{icra ="Proceedings of the IEEE Conference Robotics and Automation"}

@string{iclr ="Proceedings of International Conference on Learning and Representation"}

@string{ijcv =  "International Journal of Computer Vision"}

@string{neurips = "Advances in Neural Information Processing Systems"}

@string{CommACM = "Comm. Assoc. Comp. Mach."}

@string{ijdar = "International Journal of Document Analysis and Recognition"}

@string{icdar = "International conference on document analysis and recognition"}

@string{mar = "March"}

@inproceedings{nguyen2021dictionary,
  title={Dictionary-guided scene text recognition},
  author={Nguyen, Nguyen and Nguyen, Thu and Tran, Vinh and Tran, Minh-Triet and Ngo, Thanh Duc and Nguyen, Thien Huu and Hoai, Minh},
  booktitle=cvpr,
  year={2021}
}

@article{Nikouei_2025,
  title={Small object detection: A comprehensive survey on challenges, techniques and real-world applications},
  author={Nikouei, Mahya and Baroutian, Bita and Nabavi, Shahabedin and Taraghi, Fateme and Aghaei, Atefe and Sajedi, Ayoob and Moghaddam, Mohsen Ebrahimi},
  journal={Intelligent Systems with Applications},
  pages={200561},
  year={2025},
  publisher={Elsevier}
}

@inproceedings{ye2023deepsolo,
  title={Deepsolo: Let transformer decoder with explicit points solo for text spotting},
  author={Ye, Maoyuan and Zhang, Jing and Zhao, Shanshan and Liu, Juhua and Liu, Tongliang and Du, Bo and Tao, Dacheng},
  booktitle=cvpr,
  year={2023}
}

@article{zhang2021character,
  title={Character-level street view text spotting based on deep multisegmentation network for smarter autonomous driving},
  author={Zhang, Chongsheng and Tao, Yuefeng and Du, Kai and Ding, Weiping and Wang, Bin and Liu, Ji and Wang, Wei},
  journal={IEEE Transactions on Artificial Intelligence},
  volume={3},
  number={2},
  pages={297--308},
  year={2021},
  publisher={IEEE}
}

@article{wu2024dstext,
  title={DSText V2: A comprehensive video text spotting dataset for dense and small text},
  author={Wu, Weijia and Zhang, Yiming and He, Yefei and Zhang, Luoming and Lou, Zhenyu and Zhou, Hong and Bai, Xiang},
  journal={Pattern Recognition},
  volume={149},
  pages={110177},
  year={2024},
  publisher={Elsevier}
}

@article{lucas2005icdar,
  title={ICDAR 2003 robust reading competitions: entries, results, and future directions},
  author={Lucas, Simon M and Panaretos, Alex and Sosa, Luis and Tang, Anthony and Wong, Shirley and Young, Robert and Ashida, Kazuki and Nagai, Hiroki and Okamoto, Masayuki and Yamamoto, Hiroaki and others},
  journal={International Journal of Document Analysis and Recognition (IJDAR)},
  volume={7},
  number={2},
  pages={105--122},
  year={2005},
  publisher={Springer}
}

@inproceedings{shahab2011icdar,
  title={ICDAR 2011 robust reading competition challenge 2: Reading text in scene images},
  author={Shahab, Asif and Shafait, Faisal and Dengel, Andreas},
  booktitle=icdar,
  pages={1491--1496},
  year={2011},
  organization={IEEE}
}

@inproceedings{karatzas2013icdar,
  title={ICDAR 2013 robust reading competition},
  author={Karatzas, Dimosthenis and Shafait, Faisal and Uchida, Seiichi and Iwamura, Masakazu and i Bigorda, Lluis Gomez and Mestre, Sergi Robles and Mas, Joan and Mota, David Fernandez and Almazan, Jon Almazan and De Las Heras, Lluis Pere},
  booktitle=icdar,
  pages={1484--1493},
  year={2013},
  organization={IEEE}
}

@inproceedings{zhang2016multi,
  title={Multi-oriented text detection with fully convolutional networks},
  author={Zhang, Zheng and Zhang, Chengquan and Shen, Wei and Yao, Cong and Liu, Wenyu and Bai, Xiang},
  booktitle=cvpr,
  year={2016}
}

@inproceedings{nagy2011neocr,
  title={NEOCR: A configurable dataset for natural image text recognition},
  author={Nagy, Robert and Dicker, Anders and Meyer-Wegener, Klaus},
  booktitle={International Workshop on Camera-Based Document Analysis and Recognition},
  pages={150--163},
  year={2011},
  organization={Springer}
}

@article{zhou2015icdar,
  title={Icdar 2015 text reading in the wild competition},
  author={Zhou, Xinyu and Zhou, Shuchang and Yao, Cong and Cao, Zhimin and Yin, Qi},
  journal={arXiv preprint arXiv:1506.03184},
  year={2015}
}

@inproceedings{shi2017icdar2017,
  title={Icdar2017 competition on reading chinese text in the wild (rctw-17)},
  author={Shi, Baoguang and Yao, Cong and Liao, Minghui and Yang, Mingkun and Xu, Pei and Cui, Linyan and Belongie, Serge and Lu, Shijian and Bai, Xiang},
  booktitle=icdar,
  volume={1},
  pages={1429--1434},
  year={2017},
  organization={IEEE}
}

@inproceedings{ye2023dptext,
  title={Dptext-detr: Towards better scene text detection with dynamic points in transformer},
  author={Ye, Maoyuan and Zhang, Jing and Zhao, Shanshan and Liu, Juhua and Du, Bo and Tao, Dacheng},
  booktitle=aaai,
  year={2023}
}

@article{wu2024end,
  title={End-to-end video text spotting with transformer},
  author={Wu, Weijia and Cai, Yuanqiang and Shen, Chunhua and Zhang, Debing and Fu, Ying and Zhou, Hong and Luo, Ping},
  journal=ijcv,
  volume={132},
  number={9},
  pages={4019--4035},
  year={2024},
  publisher={Springer}
}

@article{he2024gomatching,
  title={Gomatching: A simple baseline for video text spotting via long and short term matching},
  author={He, Haibin and Ye, Maoyuan and Zhang, Jing and Liu, Juhua and Du, Bo and Tao, Dacheng},
  journal=neurips,
  volume={37},
  pages={25663--25686},
  year={2024}
}

@inproceedings{liu2024grounding,
  title={Grounding dino: Marrying dino with grounded pre-training for open-set object detection},
  author={Liu, Shilong and Zeng, Zhaoyang and Ren, Tianhe and Li, Feng and Zhang, Hao and Yang, Jie and Jiang, Qing and Li, Chunyuan and Yang, Jianwei and Su, Hang and others},
  booktitle=eccv,
  pages={38--55},
  year={2024},
  organization={Springer}
}

@inproceedings{ravi2024sam,
  title={Sam 2: Segment anything in images and videos},
  author={Ravi, Nikhila and Gabeur, Valentin and Hu, Yuan-Ting and Hu, Ronghang and Ryali, Chaitanya and Ma, Tengyu and Khedr, Haitham and R{\"a}dle, Roman and Rolland, Chloe and Gustafson, Laura and others},
  booktitle=iclr,
  volume={2025},
  pages={28085--28128},
  year={2025}
}

@article{fischler1981random,
  title={Random sample consensus: a paradigm for model fitting with applications to image analysis and automated cartography},
  author={Fischler, Martin A and Bolles, Robert C},
  journal=CommACM,
  volume={24},
  number={6},
  pages={381--395},
  year={1981},
  publisher={ACM New York, NY, USA}
}

@inproceedings{reddy2020roadtext,
  title={Roadtext-1k: Text detection \& recognition dataset for driving videos},
  author={Reddy, Sangeeth and Mathew, Minesh and Gomez, Lluis and Rusinol, Mar{\c{c}}al and Karatzas, Dimosthenis and Jawahar, CV},
  booktitle=icra,
  pages={11074--11080},
  year={2020},
  organization={IEEE}
}

@article{ch2020total,
  title={Total-text: toward orientation robustness in scene text detection},
  author={Ch’ng, Chee-Kheng and Chan, Chee Seng and Liu, Cheng-Lin},
  journal=ijdar,
  volume={23},
  number={1},
  pages={31--52},
  year={2020},
  publisher={Springer}
}

@inproceedings{bertinetto2016fully,
  title={Fully-convolutional siamese networks for object tracking},
  author={Bertinetto, Luca and Valmadre, Jack and Henriques, Joao F and Vedaldi, Andrea and Torr, Philip HS},
  booktitle=eccv,
  pages={850--865},
  year={2016},
  organization={Springer}
}

@inproceedings{li2017perceptual,
  title={Perceptual generative adversarial networks for small object detection},
  author={Li, Jianan and Liang, Xiaodan and Wei, Yunchao and Xu, Tingfa and Feng, Jiashi and Yan, Shuicheng},
  booktitle=cvpr,
  year={2017}
}

@inproceedings{gupta2016synthetic,
  title={Synthetic data for text localisation in natural images},
  author={Gupta, Ankush and Vedaldi, Andrea and Zisserman, Andrew},
  booktitle=cvpr,
  year={2016}
}

@article{wu2021bilingual,
  title={A bilingual, openworld video text dataset and end-to-end video text spotter with transformer},
  author={Wu, Weijia and Cai, Yuanqiang and Zhang, Debing and Wang, Sibo and Li, Zhuang and Li, Jiahong and Tang, Yejun and Zhou, Hong},
  journal={arXiv preprint arXiv:2112.04888},
  year={2021}
}

@article{chen2004automatic,
  title={Automatic detection and recognition of signs from natural scenes},
  author={Chen, Xilin and Yang, Jie and Zhang, Jing and Waibel, Alex},
  journal={IEEE Transactions on image processing},
  volume={13},
  number={1},
  pages={87--99},
  year={2004},
  publisher={IEEE}
}

@article{raisi2022text,
  title={Text detection \& recognition in the wild for robot localization},
  author={Raisi, Zobeir and Zelek, John},
  journal={arXiv preprint arXiv:2205.08565},
  year={2022}
}

@article{hassan2025attention,
  title={Attention based unified architecture for Arabic text detection on traffic panels to advance autonomous navigation in natural scenes},
  author={Hassan, Basma M and Gamel, Samah A and Talaat, Fatma M},
  journal={Scientific Reports},
  volume={15},
  number={1},
  pages={34138},
  year={2025},
  publisher={Nature Publishing Group UK London}
}

@inproceedings{yuan2025piftext,
  title={PIFText: Progressive Injection Fusion for Text Detection in Autonomous Driving Scenarios},
  author={Yuan, Chenyu and Zhang, Jing and Li, Jiafeng and Zhuo, Li},
  booktitle={IEEE International Conference on Deep Learning and Computer Vision},
  pages={1--6},
  year={2025},
  organization={IEEE}
}

@inproceedings{Zhang2021ByteTrackMT,
  title={Bytetrack: Multi-object tracking by associating every detection box},
  author={Zhang, Yifu and Sun, Peize and Jiang, Yi and Yu, Dongdong and Weng, Fucheng and Yuan, Zehuan and Luo, Ping and Liu, Wenyu and Wang, Xinggang},
  booktitle=eccv,
  pages={1--21},
  year={2022},
  organization={Springer}
}

@article{Chen2023SeqTrackST,
  title={SeqTrack: Sequence to Sequence Learning for Visual Object Tracking},
  author={Xin Chen and Houwen Peng and Dong Wang and Huchuan Lu and Han Hu},
  journal=cvpr,
  year={2023},
}

@inproceedings{fan2019lasot,
  title={Lasot: A high-quality benchmark for large-scale single object tracking},
  author={Fan, Heng and Lin, Liting and Yang, Fan and Chu, Peng and Deng, Ge and Yu, Sijia and Bai, Hexin and Xu, Yong and Liao, Chunyuan and Ling, Haibin},
  booktitle=cvpr,
  year={2019}
}

@inproceedings{lao2019minimum,
  title={Minimum delay object detection from video},
  author={Lao, Dong and Sundaramoorthi, Ganesh},
  booktitle=cvpr,
  year={2019}
}

@inproceedings{li2019siamrpn++,
  title={Siamrpn++: Evolution of siamese visual tracking with very deep networks},
  author={Li, Bo and Wu, Wei and Wang, Qiang and Zhang, Fangyi and Xing, Junliang and Yan, Junjie},
  booktitle=cvpr,
  year={2019}
}

@inproceedings{yan2021learning,
  title={Learning spatio-temporal transformer for visual tracking},
  author={Yan, Bin and Peng, Houwen and Fu, Jianlong and Wang, Dong and Lu, Huchuan},
  booktitle=cvpr,
  year={2021}
}

@inproceedings{cui2022mixformer,
  title={Mixformer: End-to-end tracking with iterative mixed attention},
  author={Cui, Yutao and Jiang, Cheng and Wang, Limin and Wu, Gangshan},
  booktitle=cvpr,
  year={2022}
}

@inproceedings{karaev2025cotracker3,
  title={Cotracker3: Simpler and better point tracking by pseudo-labelling real videos},
  author={Karaev, Nikita and Makarov, Yuri and Wang, Jianyuan and Neverova, Natalia and Vedaldi, Andrea and Rupprecht, Christian},
  booktitle=cvpr,
  year={2025}
}

@inproceedings{bai2024artrackv2,
  title={Artrackv2: Prompting autoregressive tracker where to look and how to describe},
  author={Bai, Yifan and Zhao, Zeyang and Gong, Yihong and Wei, Xing},
  booktitle=cvpr,
  year={2024}
}

@inproceedings{m_Zhang-Hoai-CVPR23,
 author = {Zekun Zhang and Minh Hoai},
 title = {Object Detection with Self-Supervised Scene Adaptation},
 year = {2023},
 booktitle = {Proceedings of the {IEEE} Conference on Computer Vision and Pattern Recognition (CVPR)},
 doi = {10.1109/CVPR52729.2023.02068},
}

@inproceedings{m_Zhang-etal-BMVC24,
 author = {Zekun Zhang and Vu Quang Truong and Minh Hoai},
 title = {Efficiency-preserving Scene-adaptive Object Detection},
 year = {2024},
 booktitle = {Proceedings of British Machine Vision Conference},
}

@inproceedings{m_Zhang-etal-BMVC21,
 author = {Zekun Zhang and Farrukh M. Koraishy and Minh Hoai},
 title = {Exemplar-Based Early Event Prediction in Video},
 year = {2021},
 booktitle = {Proceedings of British Machine Vision Conference},
}

@article{m_Hoai-DelaTorre-IJCV14,
 author = {Minh Hoai and Fernando {De la Torre}},
 title = {Max-Margin Early Event Detectors},
 year = {2014},
 journal = {International Journal of Computer Vision},
 volume = {107},
 number = {2},
 pages = {191--202},
 doi = {10.1007/s11263-013-0683-3},
}

@inproceedings{m_Hoai-DelaTorre-CVPR12,
 author = {Minh Hoai and Fernando {De la Torre}},
 title = {Max-Margin Early Event Detectors},
 year = {2012},
 booktitle = {Proceedings of the {IEEE} Conference on Computer Vision and Pattern Recognition (CVPR)},
 doi = {10.1109/CVPR.2012.6248012},
}

@inproceedings{m_Tran-etal-ICIP21a,
 author = {Vinh Tran and Niranjan Balasubramanian and Minh Hoai},
 title = {Progressive Knowledge Distillation for Early Action Recognition},
 year = {2021},
 booktitle = {IEEE International Conference on Image Processing},
 doi = {10.1109/ICIP42928.2021.9506507},
}

@inproceedings{m_Tran-etal-ICIP21b,
 author = {Vinh Tran and Yang Wang and Zekun Zhang and Minh Hoai},
 title = {Knowledge Distillation for Human Action Anticipation},
 year = {2021},
 booktitle = {IEEE International Conference on Image Processing},
 doi = {10.1109/ICIP42928.2021.9506693},
}

@inproceedings{lee2013pseudo,
  author    = {Dong-Hyun Lee},
  title     = {Pseudo-Label: The Simple and Efficient Semi-Supervised Learning Method for Deep Neural Networks},
  booktitle = {ICML Workshop on Challenges in Representation Learning},
  year      = {2013}
}

@inproceedings{xie2020noisy,
  author    = {Qizhe Xie and Minh{-}Thang Luong and Eduard Hovy and Quoc V. Le},
  title     = {Self-Training with Noisy Student Improves ImageNet Classification},
  booktitle = {Proceedings of the IEEE/CVF Conference on Computer Vision and Pattern Recognition (CVPR)},
  year      = {2020},  
}
\end{document}